\documentclass[letterpaper, 10 pt, conference]{ieeeconf}  

\IEEEoverridecommandlockouts                              

\usepackage{cite}
\usepackage{amsfonts}
\usepackage{algorithmic}
\usepackage{graphicx}
\usepackage{textcomp}

\usepackage{amsmath}
\usepackage{amssymb}

\usepackage{amsthm}
\usepackage{graphicx}
\usepackage{mathtools}

\usepackage[hidelinks]{hyperref}
\usepackage{colortbl}

\usepackage[linesnumbered,ruled,vlined]{algorithm2e}

\SetCommentSty{mycommfont}
\SetKwInput{KwInput}{Input}
\SetKwInput{KwOutput}{Output}

\title{\LARGE \bf
Risk-Sensitive Extended Kalman Filter
}

\author{Armand Jordana, Avadesh Meduri, Etienne Arlaud, Justin Carpentier and Ludovic Righetti
\thanks{This work was in part supported by the National Science Foundation grants 1932187, 1925079, 2026479, 2222815 and the French government under management of Agence Nationale de la Recherche as part of the ”Investissements d’avenir” program, reference ANR-19-P3IA-0001 (PRAIRIE 3IA Institute), the  European Union through the AGIMUS project (GA no.101070165) and the Louis Vuitton ENS Chair on Artificial Intelligence.}
\thanks{Armand Jordana, Avadesh Meduri and Ludovic Righetti are with the Tandon School of Engineering, New York University, Brooklyn, NY.
(e-mail: aj2988@nyu.edu, am9789@nyu.edu, lr114@nyu.edu). }
\thanks{Justin Carpentier and Etienne Arlaud are with Inria, Département d’informatique de l’ENS, \'Ecole normale supérieure, CNRS, PSL Research University, Paris, France (e-mail: justin.carpentier@inria.fr).}
}

\begin{document}

\maketitle
\thispagestyle{empty}
\pagestyle{empty}

\begin{abstract}
In robotics, designing robust algorithms in the face of estimation uncertainty is a challenging task. Indeed, controllers often do not consider the estimation uncertainty and only rely on the most likely estimated state. Consequently, sudden changes in the environment or the robot's dynamics can lead to catastrophic behaviors. In this work, we present a risk-sensitive Extended Kalman Filter that allows doing output-feedback Model Predictive Control (MPC) safely. 
This filter adapts its estimation to the control objective.
By taking a pessimistic estimate concerning the value function resulting from the MPC controller, the filter provides increased robustness to the controller in phases of uncertainty as compared to a standard Extended Kalman Filter (EKF).
Moreover, the filter has the same complexity as an EKF, so that it can be used for real-time model-predictive control. 
The paper evaluates the risk-sensitive behavior of the proposed filter when used in a nonlinear model-predictive control loop on a planar drone and industrial manipulator in simulation, as well as on an external force estimation task on a real quadruped robot.
These experiments demonstrate the abilities of the approach to improve performance in the face of uncertainties significantly.
\end{abstract}

\section{INTRODUCTION}

Adapting the decisions robots make based on their perception of the world is key to deploying robots outside factories. More precisely, controllers should adapt to the degree of certainty or confidence of the robot's belief of the world.
For instance, it is important that a quadruped chooses conservative footholds and slows body movements when its confidence in the location of the ground decreases. Robust output feedback Model Predictive Control (MPC) studies methods that can adapt robot decisions based on the confidence of the perception module. However, the general nonlinear problem is very difficult, and practical algorithms remain often limited to linear systems \cite{mayne2014model}.

The common practice in robotics is to decouple estimation and control (i.e., assume that the certainty equivalence principle holds) \cite{kuindersma2016optimization,neunert2018whole,sleiman2021unified, kim2019highly, dantec2022whole}. 
This approach is often chosen due to the availability of separate and tractable control and estimation algorithms that can be deployed on the robot. The estimation module is often a variation of a Gaussian filter, such as an Extended Kalman Filter (EKF)\,\cite{kalman}, which computes both the mean and uncertainty of the state estimates from sensor information. In control, an increasingly popular approach is MPC, which consists in solving an optimal control problem numerically, e.g., using Differential Dynamic Programming (DDP) \cite{erez2013integrated, koenemann2015whole, neunert2018whole, budhiraja2018differential, kleff2021high}, at each time step or at a fixed frequency. The controller can then adapt its behavior online based on the current robot and environment states. 
During deployment, the estimation module is used to compute the mean of the state estimate, which is then passed on to the controller to compute the optimal behavior \cite{sleiman2021unified, dantec2022whole, neunert2018whole,kim2019highly, kuindersma2016optimization}. Unfortunately, relying on the most likely outcome can lead to catastrophic behavior. For instance, on a load-carrying task with a quadruped where the load's mass is unknown, the notion of mean might not be appropriate as this could lead the quadruped to apply insufficient force on the ground and then fall. 

Some approaches try to address this issue by adding robustness or safety bounds in either the estimation or control block while keeping them independent. For instance, Robust Extended Kalman filtering \cite{einicke1999robust} adds robustness to inaccuracies of the EKF or the model. However, the control objective is disregarded, and therefore the controller cannot be robust to estimation uncertainties. 
Robust MPC has been studied and applied to robots. \cite{villa2017model} used tube-based MPC to control a biped. \cite{gazar2021stochastic} used linear stochastic MPC to account for uncertainties in bipedal walking. However, this line of work assumes the state to be known. In contrast to such approaches, we aim to link estimation and control by adding into the estimation module a notion of control performance to improve robustness to the estimation uncertainty.
%


In this work, we leverage our previous theoretical results on dynamic game control with imperfect state observation\cite{jordana2022stagewise} to introduce the Risk-Sensitive Extended Kalman Filter (RS-EKF), a novel filter that enables online risk-sensitive output feedback MPC. The RS-EKF computes state estimates robust to measurement uncertainty while taking into account the value function provided by the controller, i.e., the estimator
tailors risk reduction to the control objectives.
%
This, in turn 
enables automatic modification of robot decisions to be cautious in times of high environmental perturbation. 
Furthermore, RS-EKF has a similar computational cost to an EKF, allowing real-time deployment.
To demonstrate the ability of the filter, we use it 
together with a DDP-based online non-linear controller to perform risk-sensitive output-feedback MPC on various simulated robots, such as a quadrotor subjected to arbitrary changes in its mass, and a KuKa robot facing unforeseen environmental disturbances. Finally, we test the filter on a real quadruped robot Solo12~\cite{grimminger2020open} to perform an external force estimation and balancing task. These experiments demonstrate that the robots are more robust to perturbations with the RS-EKF algorithm than a classical EKF. To the best of our knowledge, this is the first time that a non-linear risk-sensitive output-feedback MPC controller has been deployed on a robot.


\section{BACKGROUND}

\subsection{Dynamic game output feedback MPC }

To design a controller sensitive to the risk related to estimation uncertainty, Whittle \cite{whittle1981risk} introduced a zero-sum game that aims at solving jointly the estimation and control problem. Given a history of measurements  $y_{1:t}$, a history of control inputs, $u_{0:t-1}$ and a prior on the initial state~$\hat x_0 $, we aim to find a control sequence $u_{t:T-1}$ that minimizes a given cost~$\ell$ while an opposing player aims to find the disturbances $(w_{0:T}, \gamma_{1:t})$ that maximize this cost $\ell$ minus a weighted norm of the disturbances. Such a problem is formally written as:
\begin{align}\label{eq:main_problem}
&  \min_{u_{t:T-1}} \max_{w_{0:T}} \max_{\gamma_{1:t}}  \,\,\,\,   \ell_{T}(x_{T}) + \sum_{i=0}^{T-1} \ell_i (x_{i}, u_{i})  \\
    & - \frac{1}{2\mu} \left( \omega_{0}^T  P^{-1} \omega_{0}  +  \sum_{j=1}^{t} \gamma_j^T  R_j^{-1} \gamma_j  + \sum_{j=1}^{T}   w_{j} ^ T Q_j^{-1} w_{j}   \right)\nonumber
\end{align}
\vspace{-0.2cm}
\begin{subequations}\label{eq:dynamics}
\begin{align} 
\mbox{subject to} \quad  x_{0} &= \hat{x}_0 + w_{0},  & \\
x_{j+1} &= f_j(x_j, u_j) + w_{j+1},  &   0 \leq j < T, \\
y_{j} &= h_j(x_j)+ \gamma_{j},  &  1\leq j \leq t.
\end{align}
\end{subequations}

\noindent where $\mu > 0$. $x_j$ is the state, $\omega_j$ the process disturbance, $y_j$ the measurement,$\gamma_j$ the measurement disturbance, $T$ the time horizon, $t$ the current time. The transition model $f_j$, the measurement model $h_j$, and the cost $\ell_j$ are assumed to be $\mathcal{C}^2$. The measurement uncertainty $R_j$, the process uncertainty $Q_j$ and the initial state uncertainty $P$ are positive-definite matrices. This weighted sum of the disturbances can be seen as the estimation of maximum a posteriori probability (MAP) under Gaussian assumption. Hence, $R_j$, $Q_j$ and $P$ can be thought of as the covariance matrix of Gaussian noise. 

Interestingly, this problem encompasses both formulations of control and estimation. If $t = 0$, in the limit where $\mu$ tends to zero, we find the generic OCP formulation \cite{campi1996nonlinear} which directly minimizes the cost function assuming standard deterministic dynamics.
 And, if  $ t = T $ and if we consider all costs $\ell_i$ to be null, then, \eqref{eq:main_problem} is equivalent to maximizing the estimation maximum a posteriori probability (MAP).
Here, the parameter $\mu$ is referred to as the risk-sensitive parameter and regulates how adversarial the problem is.

Whittle \cite{whittle1981risk} provided a solution to this min-max problem in the case of linear dynamics and quadratic costs. This solution can be iteratively obtained with two recursions, one on past disturbances and one on future disturbances. Recently, \cite{jordana2022stagewise} showed how this solution could be used to implement an efficient Newton's method that iteratively searches for a saddle point of the more general Problem \eqref{eq:main_problem}. Exploiting the sparsity of a problem, the proposed Newton step has a linear complexity in the time horizon (note that naive optimization has at least a cubic complexity \cite{copp2017simultaneous}). The solution can then be interpreted as a risk-sensitive Kalman smoother coupled to minimax DDP \cite{morimoto2002minimax}.

In this work, we use these insights to derive a computationally efficient risk-sensitive extended Kalman filter that can be used for output feedback MPC.
%
This is done by simplifying Problem\,\eqref{eq:main_problem} to match assumptions common to EKF and DDP while keeping the adversarial min-max formulation. 

\subsection{Extended Kalman Filter}

We now discuss the structure of the EKF necessary to derive
our filter. The EKF is usually derived by computing the probability a posteriori of the state given measurements, using the linearized dynamics and a Gaussian noise assumption \cite{thrun2002probabilistic}. However, the EKF can also be derived from an optimization point of view \cite{bell1993iterated}. More precisely, the EKF can be seen as a Gauss-Newton step around a well-chosen point on the log-likelihood of the maximum a posteriori probability (MAP), i.e.  $\log(p(x_t, x_{t-1}| y_{t}))$ \cite{bell1993iterated}.
Assuming disturbances follow Gaussian distributions, $\gamma_{t} \sim \mathcal{N}(0, R_{t})$, $\omega_{t} \sim \mathcal{N}(0, Q_{t})$, the MAP can be written as:
\begin{align}
  \max_{x_{t}, x_{t-1}} \,\,\,\,\, -(y_t - h_t(x_t) )^T  &R_t^{-1} (y_t - h_t(x_t) ) \nonumber \\
     - (x_{t} - f_{t-1}(x_{t-1}, u_{t-1}) ) ^ T &Q_t^{-1}(x_{t} - f_{t-1}(x_{t-1}, u_{t-1}) )   \nonumber \\
     -(x_{t-1} - \hat{x}_{t-1} ) ^ T &P_{t-1}^{-1} (x_{t-1} - \hat{x}_{t-1} ) \label{eq:EKF_opt}
\end{align}
\noindent where $\hat{x}_{t-1}$ is the prior knowledge on the past state and $P_{t-1}$ its associated covariance matrix.
As shown in \cite{bell1993iterated}, a Gauss-Newton step around  $\hat{x}_{t-1}$ and $\bar x_{t} = f_{t-1}(\hat x_{t-1}, u_{t-1})$ on \eqref{eq:EKF_opt} leads to well-known recursion~\cite{thrun2002probabilistic}:
\begin{align}
\bar P_{t} &= Q_{t} + F_{t-1} P_{t-1} F_{t-1}^T \label{Ppred} \\
K_t &= \bar{P}_{t}  H_{t}^T  (R_{t} + H_{t} \bar{P}_{t}  H_{t}^T)^{-1} \\
P_{t} &= (I - K_t H_{t}) \bar{P}_{t} \label{Pt1} \\
\hat \mu_{t} &=  K_t (y_{t} - h_{t}(\bar x_{t})) \label{mut1} \\
\hat x_{t} &= \bar x_{t} + \hat \mu_{t}   \label{xt1}
\end{align}
\noindent where $F_{t-1} = \partial_x f_{t-1} ( \hat x_{t-1}, u_{t-1})$ and  $H_{t} = \partial_x h_{t} ( \bar x_{t})$.
Here,  $\hat{x}_{t}$ is the most likely estimate and $P_{t}$ is the covariance uncertainty. In practice, at the next time step, $\hat{x}_{t}$ is used as the prior knowledge on the state. 

We notice the similar structure of the costs of Problem~\eqref{eq:main_problem} and Eq.\,\eqref{eq:EKF_opt},
except that the EKF only uses one measurement and does not
include the control cost $\ell_j$.
Hence, Eq.\,\eqref{eq:main_problem} can be seen as a maximization of the estimation \mbox{log-likelihood} up to some cost terms. We will leverage this similarity to derive a risk-sensitive version of the EKF. More precisely, we will add cost-dependent terms in the maximization \eqref{eq:EKF_opt} in order for the filter to adapt to the control objective.

\subsection{Nonlinear MPC}

Once the state is estimated, it can then be used in the control module. In this work, we focus on controllers defined as the solution to a minimization problem. At each time step, a stagewise cost defined over a horizon $H$ is minimized over future control inputs $u_t, u_{t+1}, \dots, u_{H-1}$:
\begin{align}
  \mathcal L_t (u_t, \dots, u_{H-1}) =  \ell_{t+H}(x_{T+H}) + \sum_{j=t}^{t+H-1} \ell_j (x_{j}, u_{j}) \label{cost}
\end{align}
\noindent where the state sequence is implicitly defined by the dynamics ${x_{j+1} = f_j(x_j, u_j)}$ and where $x_t$ is assumed to be the state estimated by the filter. At each time step, only the term $u_t$ is applied to the real system. In the next time step, the state estimate is updated given the new measurement and the optimal control problem is solved again. There are various ways to solve efficiently this minimization problem. In robotics, a popular algorithm is DDP \cite{murray1984differential} which reassembles the Newton method but with linear complexity in the time horizon. Additionally, DDP provides a quadratic approximation of the value function which we exploit in our derivation of the risk-sensitive filter.

\section{RISK SENSITIVE FILTER}
\label{section:method}
We now introduce RS-EKF, which builds on the dynamic game defined in Equation \eqref{eq:main_problem}. First, we modify the game to account for typical assumptions made for MPC while keeping the adversarial part that provides the risk-sensitive behavior. Then, we show how the solution can be computed with a Gauss-Newton step similar to the EKF, leading to an algorithm of similar complexity. Our formulation leads to a modified update in the filter of which the standard EKF can be seen as a limit case.

First, as for the EKF, we consider a history of measurements of length one. Furthermore, we disregard future uncertainties and assume deterministic dynamic equations for the future as is done in classical MPC formulations. Indeed, we expect that the high-frequency re-planning will compensate for model discrepancies.
In the end, the problem is supposed to be adversarial only with respect to the uncertainties related to the estimation. This can be written as:
\begin{align}\label{eq:main_problem_simplified}
&  \min_{u_{t:t+H-1}} \max_{w_{t}}\max_{w_{t-1}} \max_{\gamma_{t}}   \,\, \mathcal L_t (u_t, \dots, u_{H-1}) \\
    & - \frac{1}{2\mu} \left(\gamma_t^T  R_t^{-1} \gamma_t  + w_{t} ^ T Q_t^{-1} w_{t} + w_{t-1} ^ T P_{t-1}^{-1} w_{t-1}  \right)\nonumber
\end{align} 
\vspace{-0.5cm}
\begin{subequations}
\begin{align} 
\mbox{s.t.} \quad  x_{t-1} &= \hat{x}_{t-1} + w_{t-1},   \label{changeofvariable1} & \\
x_{t} &= f_{t-1}(x_{t-1}, u_{t-1}) + w_{t},  &   \label{changeofvariable2}  \\
y_{t} &= h_t(x_t)+ \gamma_{t}.&  \label{changeofvariable3} \\
x_{j+1} &= f_j(x_j, u_j),  &   t < j < T. 
\end{align}
\end{subequations}
As presented in \cite{jordana2022stagewise}, one of the key features of the dynamic game is that some of the constraints can be removed with an appropriate change of variable. Indeed, we can use the equality constraints of Equations \eqref{changeofvariable1}, \eqref{changeofvariable2} and \eqref{changeofvariable3} to replace the disturbance maximization into a maximization over $x_{t-1}, x_t$:
\begin{align}
& \min_{u_{t:t+H-1}}  \max_{x_{t-1}, x_t}  \,\, \mathcal L_t (u_t, \dots, u_{H-1})  \label{with_change_of_variable} \\
    & - \frac{1}{2\mu}(y_t - h_t(x_t) )^T  R_t^{-1} (y_t - h_t(x_t) ) \nonumber \\
    & - \frac{1}{2\mu} (x_{t} - f_{t-1}(x_{t-1}, u_{t-1}) ) ^ T Q_t^{-1}(x_{t} - f_{t-1}(x_{t-1}, u_{t-1}) )   \nonumber \\
    & - \frac{1}{2\mu} (x_{t-1} - \hat{x}_{t-1} ) ^ T P_{t-1}^{-1} (x_{t-1} - \hat{x}_{t-1} )  \nonumber \\
& \mbox{subject to} \quad x_{j+1} = f_j(x_j, u_j), \quad   t < j < T,   \nonumber
\end{align}
By definition of the MAP \cite{thrun2002probabilistic}, this can be written:
\begin{align}
\min_{u_{t:t+H-1}}   \max_{x_{t-1}, x_t} &  \mathcal L_t (u_t, \dots, u_{H-1}) - \frac{1}{\mu} \log(p(x_{t}, x_{t-1}|y_{t})) \nonumber \\
\mbox{subject to} & \quad x_{j+1} = f_j(x_j, u_j),   \quad    t < j < T.     \label{with_change_of_variable} 
\end{align}

Problem \eqref{with_change_of_variable} is intractable in the general case. However, by taking the concave-convex assumption, the minimization and maximization can be interchanged according to the minimax theorem. Consequently, the problem is equivalent~to:
\begin{align}
 \max_{x_{t-1}, x_{t}} \ \  \log(p(x_{t-1}, x_{t}|y_{t})) + \mu V_{t}(x_{t}), \label{erskf_objective}
\end{align}
\noindent where $V_t$ is the value function of the OCP:
\begin{align}
    V_t(x_t) = \min_{u_{t:t+H-1}} \mathcal L_t (u_t, \dots, u_{H-1})
\end{align}
Note that in the simplification from Eq. \eqref{eq:main_problem} to Eq. \eqref{eq:main_problem_simplified},  it is not necessary to disregard future uncertainties as the value function could be the one resulting from minimax DDP~\cite{morimoto2002minimax}.
If $\mu=0$, we will obtain the unbiased estimate of Kalman filtering and the estimate will be independent of the control objective. Otherwise, if $\mu > 0$, the term $\mu V(x_t) $ will bias the estimate towards regions with higher value function, which in turn will force the controller to be more conservative. 

We now take a Gauss-Newton step on the objective of Eq. \eqref{erskf_objective} around the prior: $\hat{x}_{t-1}$ and $\bar x_{t} = f_{t-1}(\hat x_{t-1}, u_{t-1})$.
$V_{t}(x_{t})$ is independent of $x_{t-1}$ therefore, as shown in the appendix, the maximization over $x_{t-1}$ can be simplified to:
\begin{align}
\max_{x_{t}} & - \frac{1}{2} (x_{t} - \hat{x}_{t})^T P_{t}^{-1} (x_{t} - \hat{x}_{t})  \\
& + \mu \frac{1}{2} (x_{t} - \bar x_{t})^T V^{xx}_{t} (x_{t} - \bar x_{t}) + \mu  (x_{t} - \bar x_{t})^T v_t^x \nonumber
\end{align}
where $\hat{x}_{t}$ and  $ P_{t}$ are defined as in Eq. \eqref{xt1} and \eqref{Pt1}.
where $V^{xx}_{t}$ (respectively $v^{x}_{t}$) is the hessian (respectively the gradient) of the value function. Those are typically provided by optimal control algorithms such as DDP. In the end, the solution on the maximization over $x_t$ is:
\begin{align}
\hat x_{t}^{RS} = \bar x_{t} +  (I - \mu P_{t} V^{xx}_{t})^{-1} ( \hat \mu_{t} + \mu P_{t} v^{x}_{t}) \label{rs_update}
\end{align}

Interestingly, if $\mu = 0$, we recover the EKF. This was to be expected as, when $\mu$ tends to zero, the solution of problem \eqref{eq:main_problem} is exactly the solution of the neutral case where estimation and control are solved independently\cite{whittle1981risk}.
Otherwise, the estimate is shifted towards regions with  higher cost values. 
Importantly, the magnitude of the shift depends on $P_t$ the covariance matrix of the estimation.
Note that $\mu$ cannot be arbitrarily large as $(I - \mu P_{t+1} V^{xx}_{t+1})$ needs to be positive definite. Larger values of $\mu$ would make the min-max problem defined in Eq. \eqref{eq:main_problem} ill-posed. More details on this limit value can be found in \cite{whittle1981risk}. 
In the end, the estimate is shifted towards $P_t v^{x}_t$, i.e. towards a region with a larger cost function, and the magnitude of this shift is increased in the direction corresponding to large eigenvalues of $P_{t} V^{xx}_{t}$.

We obtained the solution to the maximization problem~\eqref{with_change_of_variable}. 
Therefore, the cost function can now be minimized with respect to the control inputs by taking $\hat x_{t}^{RS}$ as an initial condition of the optimal control problem,  which can be solved for example with DDP.

Algorithm \ref{algo} summarizes the estimation procedure. It can then be used to do output-feedback MPC efficiently. At each time step, given a measurement, past control input, and a quadratic approximation of the value function, a risk-sensitive estimate can be computed. This estimate is then used to minimize the cost function \eqref{cost} for MPC and the first control input is applied to the real system. Lastly, the quadratic approximation of the value function at $t+1$ is saved as it will be used at the next estimation step.
\vspace{-0.2cm}
\begin{algorithm}[!ht]
\DontPrintSemicolon
\KwInput{$\hat x_{t-1}, u_{t-1}, y_{t}, P_{t-1}, Q_{t}, R_{t}, V_{t}, v_{t}$}
\tcc{Predict}
 $\bar P_t \gets Q_{t} + F_{t-1} P_{t-1} F_{t-1}^T$\;
 $\bar x_{t} \gets f(\hat x_{t-1}, u_{t-1})$\;
\tcc{Classical Update}
$K_t \gets \bar{P}_t  H_{t}^T  (R_{t} + H_t \bar{P}_t  H_T^T)^{-1}$\;
$P_{t} \gets (I - K_t H_{t}) \bar{P_t}$ \;
$\hat \mu_{t} \gets  K_t (y_{t} - h_{t}(\bar x_{t}))$ \;
\tcc{Value function bias}
$  p_{x_{t}} \gets (I - \mu P_{t} V^{x}_{t})^{-1} ( \hat \mu_{t} + \mu P_{t} v^{x}_{t})  $\;
$\hat x^{RS}_{t} \gets \bar x_{t}  + p_{x_{t}} $  \;
\KwOutput{$\hat x^{RS}_{t}, P_{t}$}
\caption{Risk Sensitive EKF}
\label{algo}
\end{algorithm}
\vspace{-0.2cm}

\section{EXPERIMENTS}

This section presents simulation and real robot experiments to illustrate the benefits of the proposed algorithm and demonstrate its applicability to real problems.
We study three test problems where we deploy the RS-EKF inside a MPC loop: a planar quadrotor with a load estimation task where we demonstrate qualitatively that the RS-EKF can bring conservatism appropriately in phases of high uncertainty, a push-recovery experiment on a 7-dof industrial manipulator on which we perform a quantitative study and lastly, an external force estimation task on a real quadruped robot to showcase the viability of the method on real systems.
In each experiment, we use the DDP implementation provided by Crocoddyl \cite{mastalli2020crocoddyl} to solve the optimal control problem given the filter estimate. All the code is available online\footnote{\scriptsize \url{https://github.com/machines-in-motion/risk-sensitive-EKF}}. 

\subsection{Planar quadrotor}

In this first scenario, we consider a planar quadrotor executing a load-carrying task. The goal is to move the quadrotor from position $(p_x, p_y) = (0, 0)$ to position $(1, 0)$ while carrying a load during the first half of the itinerary. The robot mass is $2$ kg and the mass's load which is unknown a priori is $3$\,kg. The system dynamics is:
\begin{align}
m \ddot p_x &= - ( u_1 + u_2) \sin(\theta) ,\nonumber\\
m \ddot p_y &=  ( u_1 + u_2) \cos(\theta) - m g, \nonumber\\
md \ddot \theta &= r ( u_1 - u_2),
\end{align}
where $m$ is the mass of the robot, $d$ the distance between the rotors, $\theta$ the orientation of the quadrotor. $u_1$ and $u_2$ are the control inputs representing the force applied at each rotor. 

In this experiment, we want to estimate online the mass parameter that changes in the middle of the flying phase. As it is standard in parameter identification \cite{voss2004nonlinear}, we augment the system's state with the unknown parameter and let it be estimated recursively by the filter (EKF or RS-EKF). The state of the system is thus:
$x = \begin{pmatrix} p_x & p_y & \theta & \dot p_x & \dot p_y & \dot \theta & m \end{pmatrix}^T$ 
\noindent and it is assumed that $ \dot m = 0 $ up to some random Gaussian noise.
The dynamics are integrated with an Euler scheme and a time step of $0.05$.
We consider that $P_0 = 10^{-4} I_7$, $R = 10^{-4} I_3$, and $Q$ is a $7 \times 7$ diagonal matrix where all terms are equal to $10^{-4}$ except the last one that we set to $2$ to represent the uncertainty in the changes of the load.
Lastly, we set $\mu = 4 \times 10^{-3}$.
A stationary cost function of the following form is considered:
\begin{align}
  &   \ell (x, u) = \alpha_1 \left(\| p_x - p_x^{des}\|^2 +  \| p_y - p_y^{des}\|^2 \right) + \alpha_2 \| \theta \|^2  \nonumber \\
    & + \alpha_3 \left( \| \dot p_x \|^2 + \| \dot p_y\|^2 + \| \dot \theta \|^2 \right) +  \alpha_4 \| u - \bar u \|^2 
\end{align}
where $\bar u = \begin{pmatrix} \frac{m g}{2}, \frac{m g}{2}  \end{pmatrix}^T$ and where: $\alpha_1 = 100$, $\alpha_2 = 10$, $\alpha_3 = 0.01$ and $\alpha_4 = 0.1 $. We consider a horizon of 20 nodes and re-plan at each new measurement, i.e. every $0.05s$.
Furthermore, we only measure: ${y = \begin{pmatrix} p_x & p_y & \theta \end{pmatrix}^T}$ to illustrate the estimator capabilities.
We simulate $4s$ with both output-feedback MPC controllers: the first one relying on the standard EKF estimate and the second one relying on the RS-EKF estimate. 

\begin{figure}[ht!]
    \centering
    \includegraphics[width=0.87\linewidth]{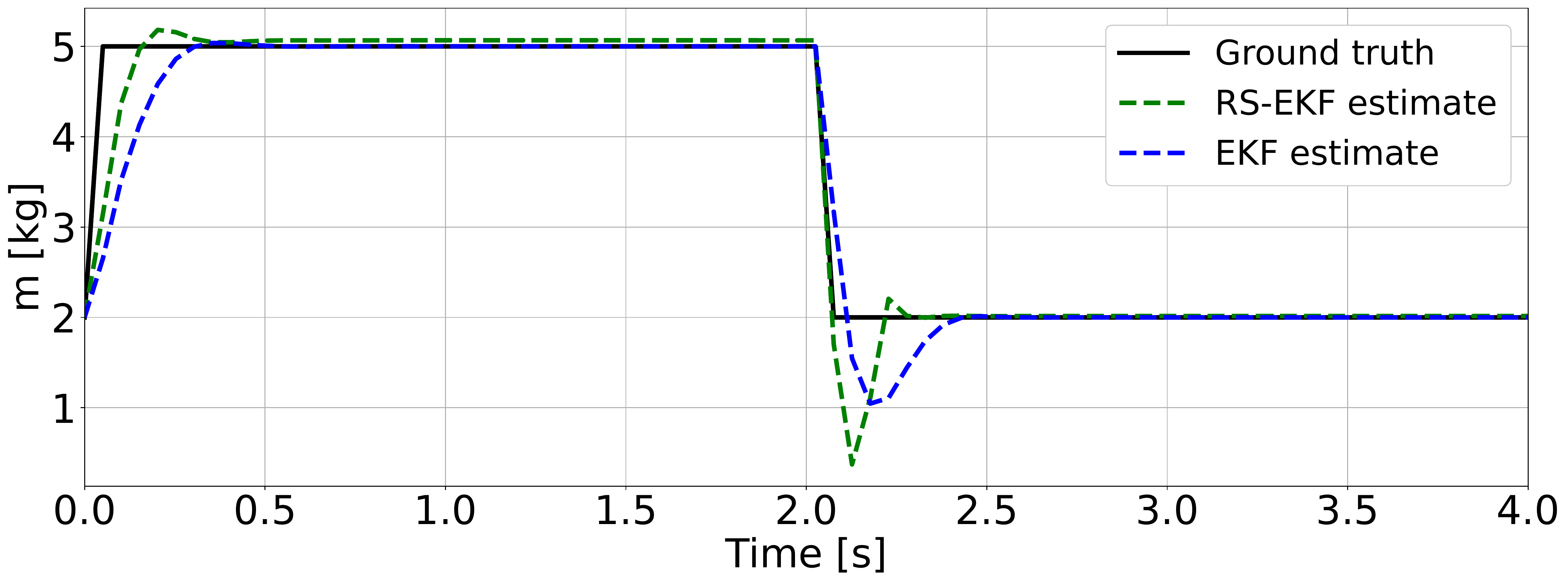}
    \caption{Mass estimation for both EKF and RS-EKF.}
    \label{quad:mass}
\end{figure}

Figure \ref{quad:mass} shows the real mass variation and the estimates of both methods. It can be seen that the RS-EKF is more reactive when the load is added or dropped. The RS-EKF estimate spikes in the phases of uncertainty, which adds some robustness. Those spikes can be explained by the fact that the uncertainty increases on the components of the state that are important in the cost function. In other words, some of the eigenvalue of $P_t V^{xx}_t$ become larger in the phases of uncertainty, which augments the shift on the estimate as shown in Equation \eqref{rs_update}.
\vspace{-0.1cm}

\begin{figure}[ht!]
    \centering
    \includegraphics[width=0.87\linewidth]{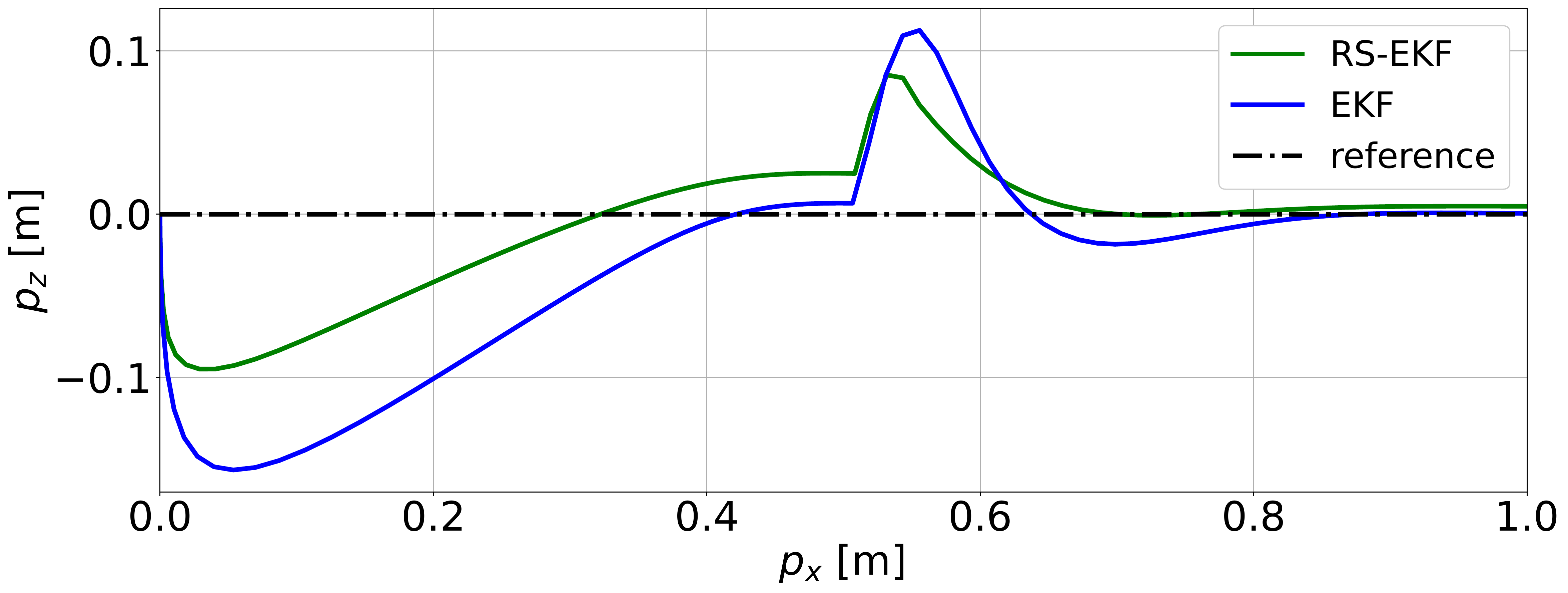}
    \caption{Quadrotor trajectory for both the EKF-MPC and the RS-EKF-MPC.}
    \label{quad:traj}
\end{figure}

Figure \ref{quad:traj} depicts the trajectory in space for both methods. It can be seen that the controller relying on the risk-sensitive estimate is more reactive. To evaluate both controllers, we compute the average Mean Square Error (MSE) relative to the reference trajectory. The MSE of RS-EKF is $0.0011$, and the one of EKF  is $0.0024$. Hence, in that situation, the risk sensitivity brings a $54 \% $ improvement in tracking. Furthermore, the average cost along the trajectory is equal to $0.0569$ for the RS-EKF-based controller, while it is equal to $ 0.0880$ for the EKF-based controller, yielding a $35\%$ improvement. This illustrates how a filter informed of the cost objective can improve the controller's performance.
\vspace{-0.1cm}

As it can be seen in both Figures \ref{quad:mass} and \ref{quad:traj}, RS-EKF introduces a steady state error. This is because the estimation uncertainty never goes to zero due to the nonzero diagonal term of the matrices $Q$ and $R$. Intuitively, it makes sense that the controller that plans for the worst is slightly sub-optimal in the phase with no environment perturbation uncertainty. The risk-sensitive filter demonstrates its benefits when there are perturbations in the environment as this corresponds to phases where the most likely estimate might be far from reality. 
In the end, this example illustrates the advantages of having a risk-sensitive controller which is conservative only in phases with large environmental perturbations.

\subsection{Kuka robot}

In this example, we consider the 7-DoF torque-controlled KUKA LWR iiwa R820 14. The 14-dimensional state is composed of the joint positions and  velocities. 
The control input is a 7-dimensional vector of the torque applied on each joint. The continuous dynamics and its analytical derivatives are provided by Pinocchio~\cite{carpentier2019pinocchio}. 
The goal is to track a reference trajectory with the end effector. To do so, we consider the following cost:
\begin{align}
    \ell_k(x_k, u_k) &= 10^{-2}  \Vert x_k - \bar x \Vert_2^2 + 10^{-4} \Vert u_k - \bar u(x_k) \Vert_2^2  \nonumber\\
    &+ 10^{2} \Vert p_k^{\text{target}} - \bar p (x_k) \Vert_2^2  \\
    \ell_T(x_T) &=  10^{2} \Vert p_T^{\text{target}} - \bar p (x_k) \Vert_2^2 + 10^{-2}  \Vert x_k - \bar x \Vert_2^2 , \nonumber 
\end{align}
$\bar x $, the initial state,  is used for the state regularization and is the concatenation of the initial configuration  $ \begin{pmatrix} 0.1 & 0.7& 0& -1& -0.5& 1.5& 0 \end{pmatrix}^T$ and a 7-dimensional zero vector corresponding to the velocity.
$\bar u(x_k)$ is the gravity compensation term given by the rigid body dynamics. $\bar p (x_k)$ is the end-effector position obtained through forward kinematics. 
$ p_k^{\text{target}}$ is defined such that the end effector follows trajectories forming a circle in the $xy$ plane. 
We use a horizon of $20$ collocation point with an integration step of $0.05s$
and re-plan at 500~Hz. 

In this experiment, we aim to showcase that the risk-sensitive filter can bring conservatism on phases with large perturbations from the environment which we simulate with large forces applied on the end effector. 
For the measurement model, we assume that all the states are observed with high accuracy, therefore, we set ${R = P_0 = 10^{-6} I_{14}}$. However, to model the disturbances in the dynamics, we set ${Q = 10^{-1} I_{14}}$. Finally, we consider $\mu = 7.5 \times 10^4$.

\begin{figure}[ht!]
    \centering
    \includegraphics[width=0.92\linewidth]{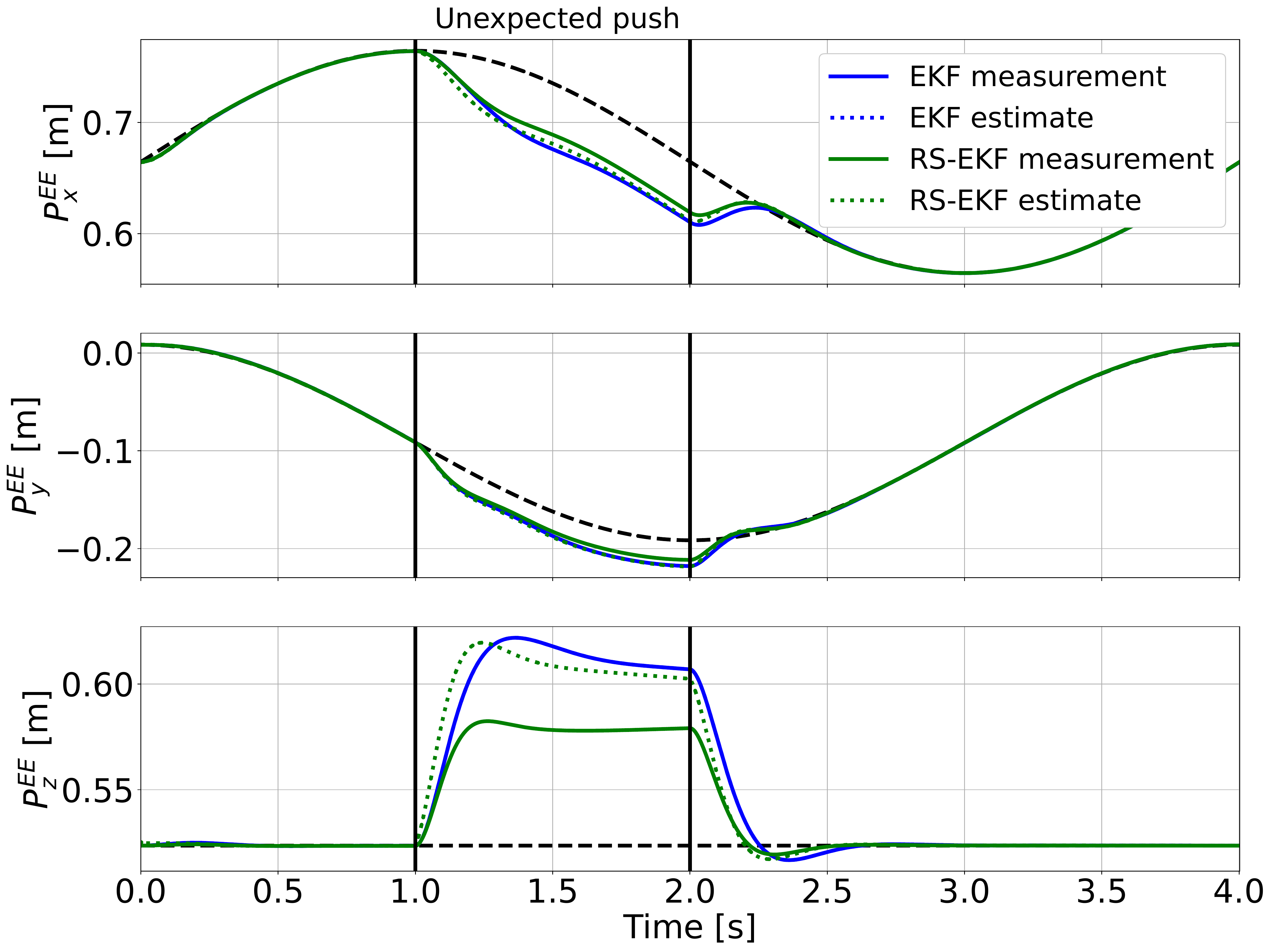}
    \vspace{-0.1cm}
    \caption{End effector trajectory on a tracking task for both the EKF-MPC and the RS-EKF-MPC. An unexpected force is applied between $1s$ and $2s$. }
    \label{fig:kukaee}
\end{figure}
Figure \ref{fig:kukaee} depicts the end effector trajectory for both controllers and their respective estimates. 
From time $1s$ to $2s$, an external force of norm $80$ is applied on the end-effector on the $x$ and $z$ direction. Both controllers are pushed away from the reference trajectory. However, the risk-sensitive estimate overestimates the distance between the reference and the end-effector.
Consequently, the robot is more aggressive in its response, and the end-effector remains closer to the reference. 
This illustrates how taking a pessimistic estimate with respect to the cost improves the running cost.
Note that both estimates are originally state estimates but are mapped to end effector space through the forward kinematics to draw Figure \ref{fig:kukaee}. The fact that those estimates are pessimistic with respect to the task originally defined in end effector space illustrates well how the method can handle nonlinear dynamics and cost functions.
\begin{figure}[ht!]
    \centering
    \includegraphics[width=0.92\linewidth]{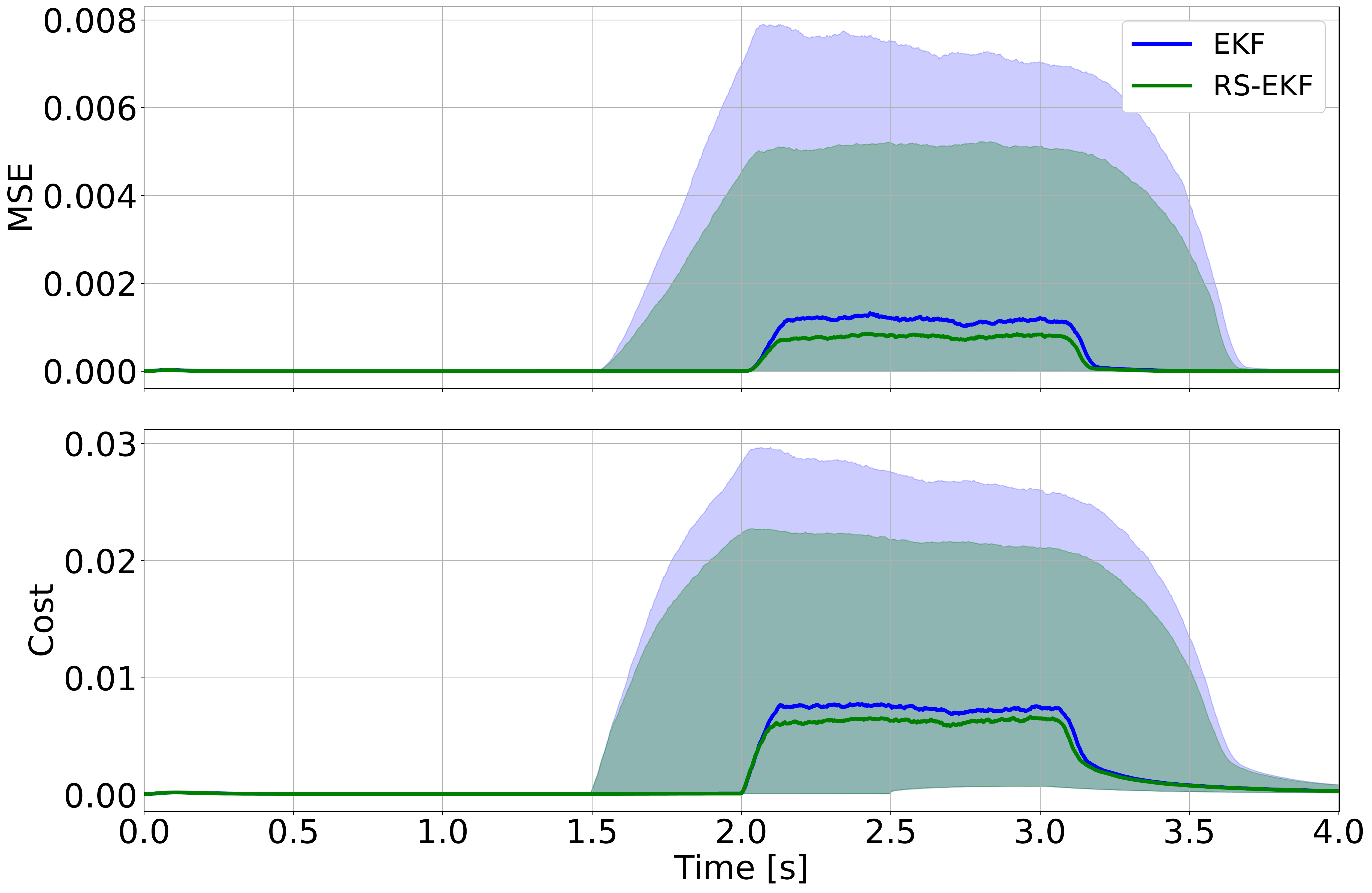}
    \vspace{-0.1cm}
    \caption{Median MSE on the tracking over $10,000$ experiments with random external disturbances. The envelope represents the $25$th and $75$th percentiles.}
    \label{fig:kuka}
\end{figure}

To validate the consistency of the filter, $10,000$ experiments are performed with random external forces. The timing and direction of the forces are uniformly sampled while the duration is fixed to $1s$ and the norm is fixed to~$80$.
Figure~\ref{fig:kuka} shows the median end effector error trajectory. In average, we obtain a $32 \%$ improvement in the MSE.
Furthermore, the mean cost is $22\%$ lower with the risk-sensitive filter. 
\begin{figure*}
    \centering
    \hspace*{0.9cm}\includegraphics[width=0.94\textwidth]{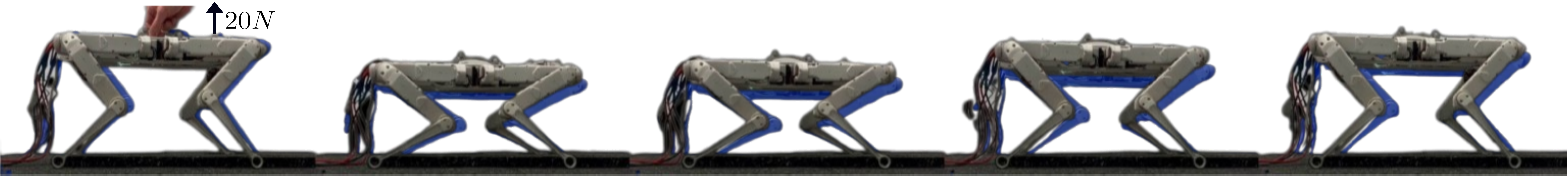} \hspace*{-0.9cm}\\
    \vspace{0.3cm}
    \includegraphics[width=0.98\textwidth]{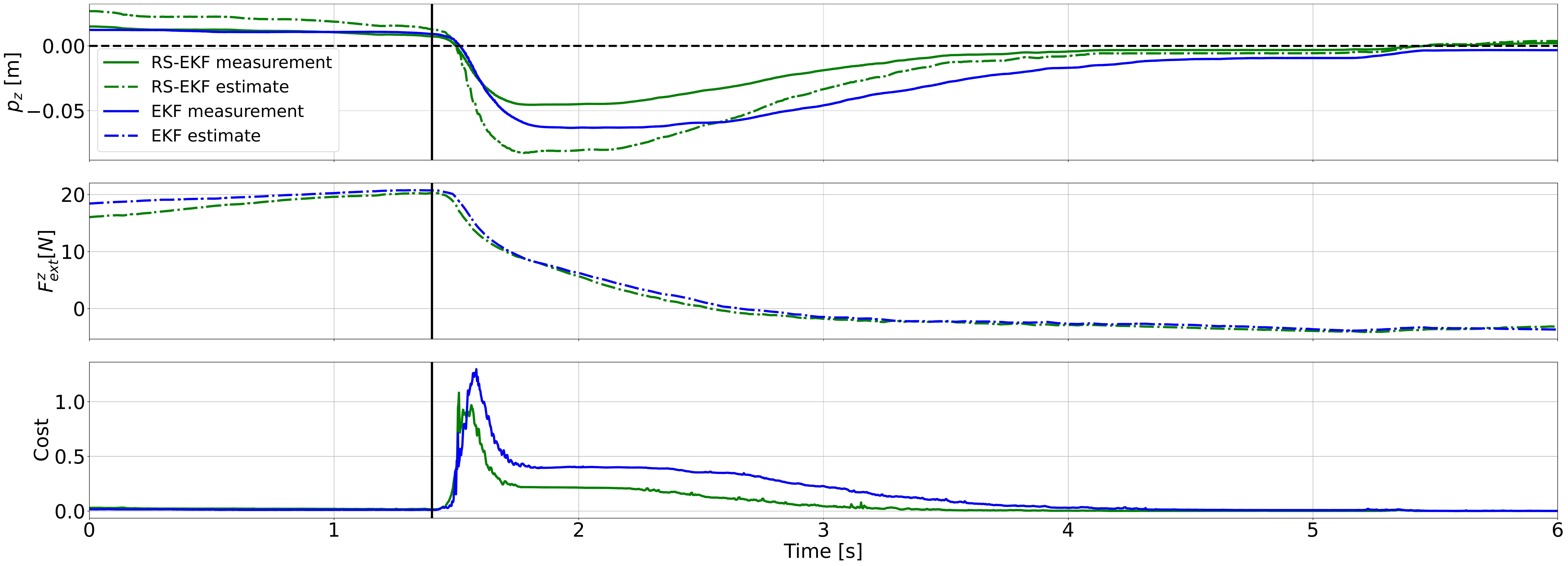}
    \vspace{-0.1cm}
    \caption{Comparison of both methods after an external force of $20$\,N is applied by pulling the robot vertically. The vertical line indicates the moment when the robot is dropped. The top sub-figure overlays the trajectory of both the EKF in blue and RS-EKF in solid.  The RS-EKF-based controller is more reactive to the perturbation and returns to the reference sooner.}
    \label{fig:hand}
    \vspace{-0.2cm}
\end{figure*}


\subsection{Load estimation on a quadruped robot}

In this experiment, we deploy the RS-EKF on a real 12-degree-of-freedom, torque-controlled quadruped robot - Solo12 \cite{grimminger2020open}. We demonstrate the superior performance of the RS-EKF in estimating wrenches applied on Solo12 while it is standing. We generate the standing behavior on Solo12 using a non-linear MPC scheme. At each control cycle, we minimize a cost function using a centroidal model to compute the optimal forces and trajectory that keep the robot's base at a desired height and orientation. Additionally, we use an augmented state to estimate external forces-torques applied to the robot \cite{rotella2015humanoid}:
\begin{align}
    \dot c &= \frac{1}{m} l \\
    \dot l &= mg + \sum_{i=1}^{M_c} F_i +  F_{\text{ext}} \\
    \dot k &= \sum_{i=1}^{M_c} (p_i-c) \times F_i +  \tau_{\text{ext}} \\
    &\dot { F}_{\text{ext}} = 0, \quad \dot {\tau}_{\text{ext}} = 0
\end{align}
\noindent where $m$ denotes the mass, $M_c$ the number of end effectors in contact. Each $p_i$ represents a contact location. Here, the state is $x = \begin{pmatrix} c & l & k &  F_{\text{ext}} &  \tau_{\text{ext}} \end{pmatrix}^T$ which includes the Center of Mass ($\textit{c}$), linear momentum ($\textit{l}$), Angular Momentum ($\textit{k}$) and External Force-torques (${F_{\text{ext}}}, \tau_{\text{ext}}$). The measurement is $y =  \begin{pmatrix} c & l & k \end{pmatrix}^T$ up to some noise.
A motion capture system measures the base position, velocity, and orientation and an IMU gives the orientation velocity. Joint encodings are provided and their velocities are derived with finite differences. Then, given, $q, \dot q$, we can compute $c, l, k$ with Pinocchio~\cite{carpentier2019pinocchio,pinocchioweb} and can then be used as a measurement by the centroidal-based filter.

The control input, $u = \begin{pmatrix} F_1  & \dots & F_{M_c} \end{pmatrix}^T$,  is a $M_c\times3$ dimensional vector, representing the force applied at each end effector. For this experiment, the robot is standing, therefore, $M_c=4$.
The cost function for the OCP is:
\begin{align}
    \ell_t (x, u) &=  (x - x^{\star})^T H_x (x - x^{\star}) + (u - u^{\star})^T H_u (u - u^{\star}) \nonumber \\
    & + 10^5 \sum_{i=1}^{M_c} \ell_{\text{barrier}} ({u}_{3 i}) \nonumber \\
    \ell_T(x) &= (x - x^{\star})^T H_x  (x - x^{\star})
\end{align}
where $H_x=\operatorname{BlockDiag}(10^2 I_3, 10 I_6)$ and $H_u$ is a diagonal matrix where the diagonal terms are made of $M_c$ times the following sequence $(10^{-4}, 10^{-4}, 10^{-6})$. Lastly,
\begin{align}
    \ell_{\text{barrier}} (u) = \left \{ \begin{array}{cc}
        u^2  & \mbox{ if } u < 0 \\
         0 & \mbox{ if } 0 < u < 10\\
         (u-10)^2 & \mbox{ if } u > 10
    \end{array}\right .
\end{align} 
Here, $x^{\star} $ is designed to keep a constant height above the ground and to keep the base horizontal, $u^{\star} $ is the gravity compensation. The reference desired angular momentum for the OCP is adapted to bring the base back to a horizontal position. We do this by computing $ k^{*} = \frac{1}{T} log_{3}(R_{t} R_{\text{des}}^T )$, where $R_{t}$ is the current base rotation matrix, $R_{\text{des}}$ the rotation matrix corresponding to the quaternion $q_{\textit{des}} = [0,0,0,1]$ and  $T$ the horizon length. The $log_{3}$ is a mapping from $SE(3)$ to $\mathfrak{se}(3)$. The nominal angular momentum aims to bring the base back to the desired orientation over the time horizon~\cite{meduri2023biconmp}. The $\ell_{\text{barrier}}$ is a quadratic barrier function that creates a soft constraint on the maximum forces the robot can apply on the ground. 

We solve this OCP at 100 Hz using Croccodyl~\cite{mastalli2020crocoddyl} and then track the desired forces using a task space impedance dynamics QP  \cite{herzog2016momentum} that we solve at 1\,kHz using \cite{bambade2022prox}.
\begin{align}
\min_{f, \tau, a}  & \quad \frac{1}{2}\,\|f - F\|^2 \\ 
\mbox{subject to}  & \quad Ma + g = J^{T}f + S^T \tau + S^T f_{\text{friction}} \nonumber \\
& \quad Ja = - \dot{J} \dot q, \nonumber
\end{align}
where $J$ is the robot contact Jacobian, $M$ is the mass matrix, $S$ is the selection matrix that projects on the actuated joints, and $g$ is the robot gravity vector at the current time step. The static friction in each joint was estimated independently and is approximately equal to $0.07$. However, to keep a continuous model, we consider 
$ {f_{\text{friction}} =  -0.07 \frac{2}{\pi} \operatorname{arctan}\left(2 S \dot q \right) }$.
The two constraints ensure dynamics satisfaction and model that the end effectors do not move. 
We update our state estimate (including the external force torque) using the RS-EKF at 200 Hz with $\mu=6$. We also use the EKF in place of the RS-EKF to compare the performance of a risk-sensitive filter while keeping the update frequency the same.   
For both filter, we consider the following parameters: $P_0 = Q = \operatorname{BlockDiag}( 10^{-3} I_6, 10^{-4} I_3, 10^{-1} I_3 , 10^{-2} I_3 )$
$R = \operatorname{BlockDiag}(10^{-4} I_3, 10^{-2} I_3 , 10^{-4} I_3 )$.

The first experiment ran on the robot is shown in Fig.~\ref{fig:hand}. Here, the base of Solo12 is pulled up (in the $z$ direction) until an external estimate of 20N is computed by both the filters (Shown in bottom sub-figure \ref{fig:hand} at time $1.4s$ with a vertical line). The base is then released to let Solo12 recover and bring its base back to the nominal desired height. The top sub-figure in Fig \ref{fig:hand} shows an overlayed pictorial comparison of Solo12 when the two filters (RS-EKF and EKF) are used on the robot. Each frame corresponds to the state of Solo12 at the same time. The opaque Solo12 image shows the state with RS-EKF, and the blue one shows the EKF. This experiment shows that the RS-EKF helps the OCP to react quicker and bring the base to the nominal location sooner. This happens because the RS-EKF underestimates the base height in $z$ as compared to EKF, which makes the OCP generate higher ground reaction forces to bring the base up sooner. In the end, both filter estimate similar external forces. As it can be seen in Fig.~\ref{fig:hand}, the external vertical force does not converge exactly to zero. We find experimentally that this estimated force is due to friction. Lastly, as it can be seen, the cost of the RS-EKF-based controller is lower after the robot is dropped. The average cost of the RS-EKF is $  0.065$ while the one of the EKF is $0.130$, which corresponds to a $50\%$ improvement. This demonstrates how a filter aware of the cost objective can improve the overall performance.

\begin{figure}[ht!]
    \centering
    \includegraphics[width=\linewidth]{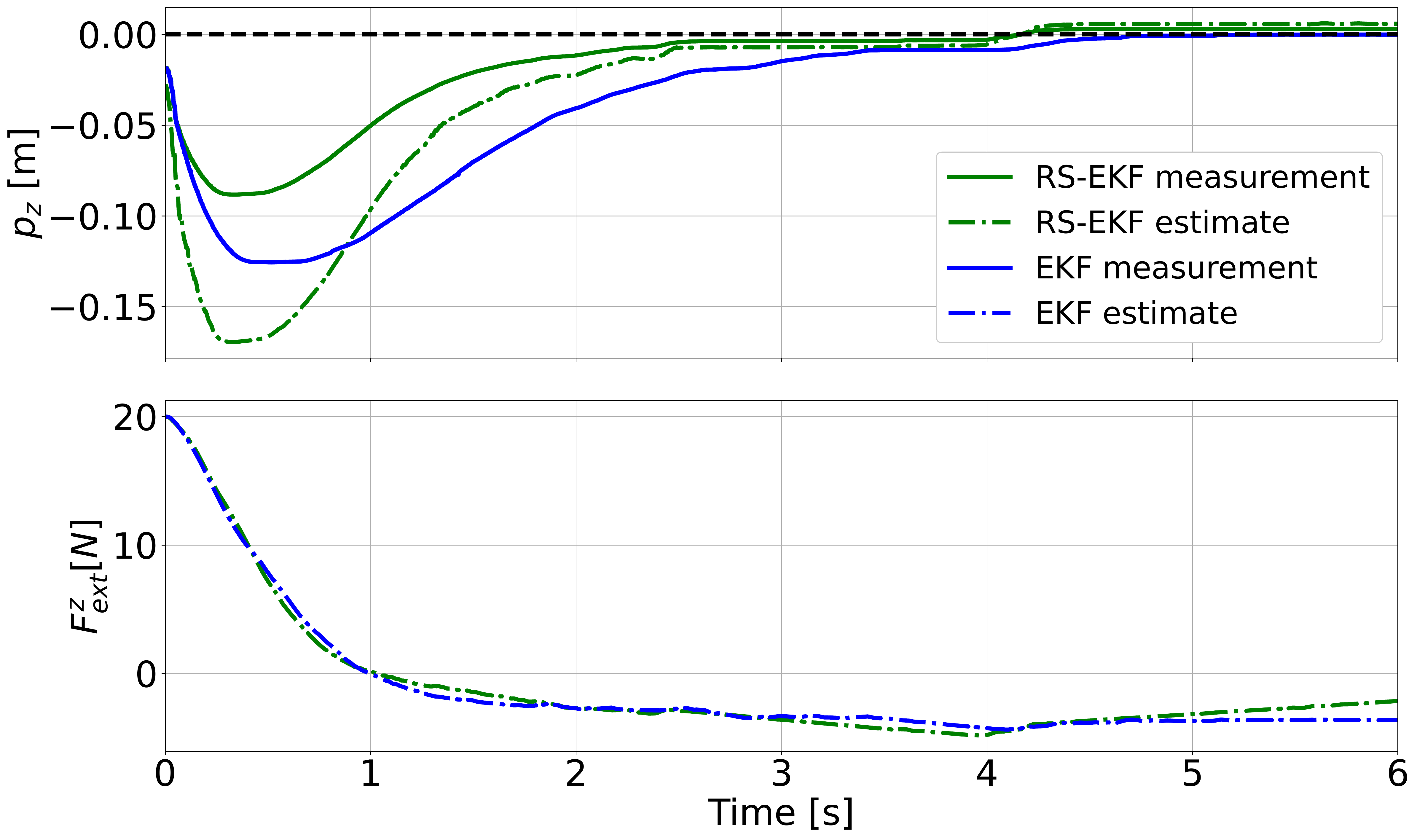}
    \caption{Comparison of the RS-EKF and EKF when initialized with a wrong prior of $20$\,N on the estimated vertical external force.}
    \label{fig:remove_block}
\end{figure}

In order to get a more systematic comparison between the filters and get rid of the human error, we perform two additional experiments where the filters are initialized with exactly the same priors. 
First, we initialize both the filters with a wrong prior on the external vertical force of $20$\,N, while in reality, no force is applied on the robot. This experiment creates an identical situation as the previous experiment while also ensuring the very same initial conditions for the robot. 
The results are shown in Fig \ref{fig:remove_block}, where the RS-EKF still performs better. Also, the performance is similar to the first experiment. In that experiment, we obtain a $ 62.9 \% $ improvement in the average cost.

\begin{figure}[ht!]
    \centering
    \includegraphics[width=\linewidth]{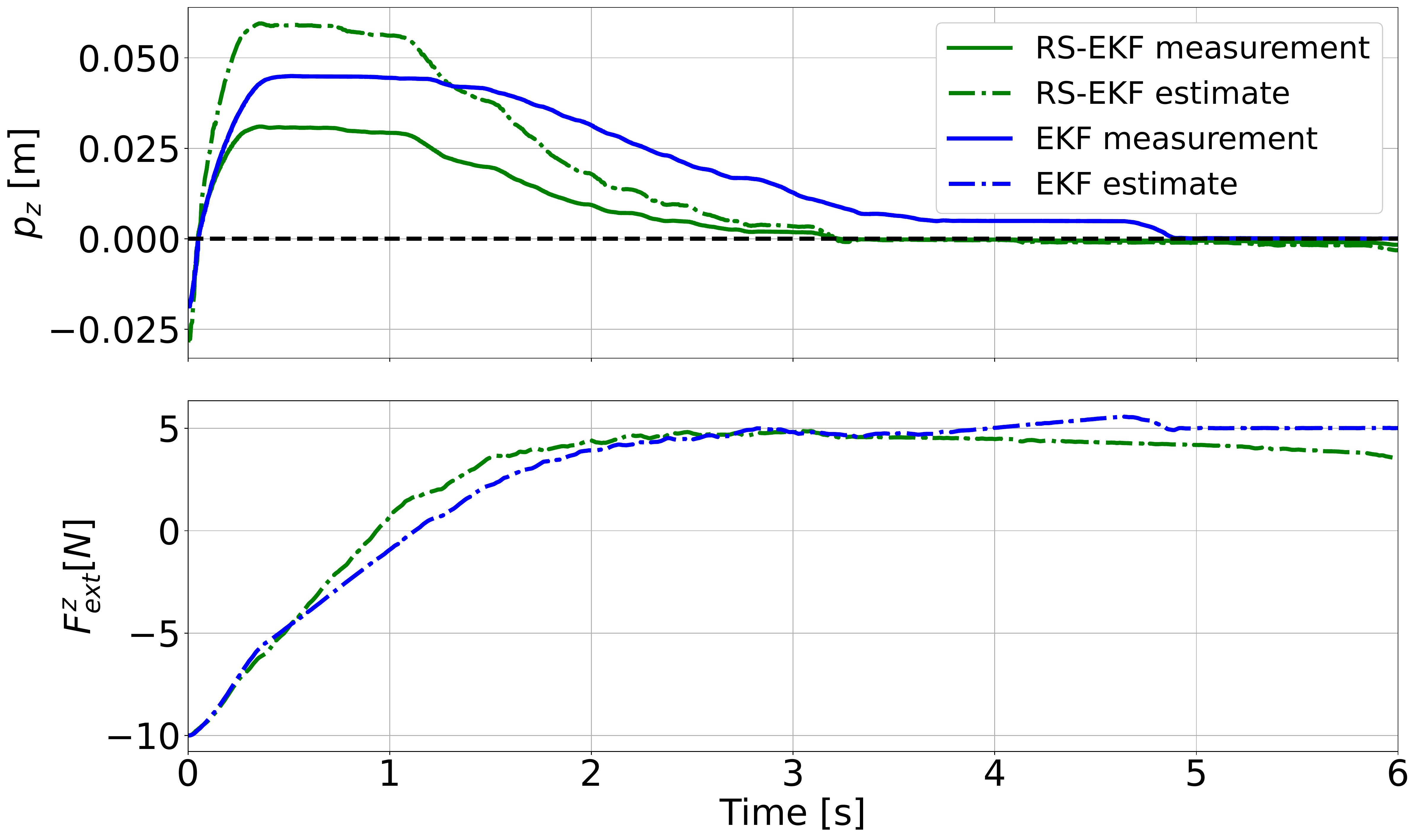}
    \caption{Comparison of the RS-EKF and EKF when initialized with a wrong prior of $-10$\, N on the estimated vertical external force.}
    \label{fig:add_block}
    \vspace{-0.4cm}
\end{figure}

Finally, we replicate the previous experiment but now, initialize the filters with a wrong prior of $-10$\, N on the external force, while, in fact, there is no force on the robot. The RS-EKF reacts sooner than EKF once again. It brings the base of Solo12 back to the desired location sooner than EKF, as can be seen in Fig\,\ref{fig:add_block}. In that experiment, we obtain a $58.9 \% $ improvement in the average cost.

\section{CONCLUSION}
To conclude, we have introduced a risk-sensitive variation of the EKF based on the zero-sum game introduced by Whittle~\cite{whittle1981risk}. 
The filter biases estimates towards high regions of
the control cost which result in more robust controllers.
Furthermore, the complexity of this filter is similar to the EKF. Lastly, we have demonstrated on several robotics problems, both in simulation and on real hardware, the benefits of this filter for output-feedback MPC. Importantly, we have shown that the proposed filter makes the controller more conservative in phases of high uncertainty leading to better overall control performance.
Future work will address scenarios involving more dynamic tasks and more complex measurement uncertainty.

\section*{APPENDIX}
By taking a quadratic approximation of the value function, the Gauss-Newton step can be written as:
\begin{align}
 &     \max_{x_{t-1}} \max_{x_t} \ \ \mu (x_{t} - \bar x_{t})^T V^{xx}_{t} (x_{t} - \bar x_{t}) + 2 \mu  (x_{t} - \bar x_{t})^T v_t^x  \nonumber\\
&-(\Delta y - H_t  \Delta x_{t} )^T R_t^{-1} (\Delta y  - H_t  \Delta x_{t} ) - \Delta x_{t-1}^T P_{t-1}^{-1} \Delta x_{t-1} \nonumber\\
 &-  ( \Delta x_{t}  - F_{t-1} \Delta x_{t-1})^T Q_t^{-1} ( \Delta x_{t}  - F_{t-1} \Delta x_{t-1}))
\end{align}
\noindent where $\Delta y = y_t -  h(\hat x_t)$, $\Delta x_{t-1} = x_{t-1} - \hat x_{t-1}$, $\Delta x_{t} = x_{t} - \hat x_{t}$.
It can then be found that $x_{t-1} = \Tilde{Q}^{-1} \Tilde{q}$,
where:
\begin{align}
    \Tilde{Q} &= P_{t-1}^{-1} + F_{t-1}^T Q_t^{-1} F_{t-1} \\
    \Tilde{q} &=  - P_{t-1}^{-1} \hat{x}_{t-1} - F_{t-1}^T Q_t^{-1} (x_t - \hat x_t) - F_{t-1}^T Q_t^{-1} F_{t-1} \hat x_{t-1}  \nonumber 
\end{align}
by using the Woodbury lemma \cite{thrun2002probabilistic},  It can be shown that:
\begin{align}
 & \max_{x_t} \ \  \mu \frac{1}{2} (x_{t} - \bar x_{t})^T V^{xx}_{t} (x_{t} - \bar x_{t}) + \mu  (x_{t} - \bar x_{t})^T v_t^x  \nonumber \\
    &- \frac{1}{2}(\Delta y - H_t \Delta x_{t})^T R_t^{-1} (\Delta y - H_t \Delta x_{t}) - \frac{1}{2} \Delta x_{t}^T \bar P_t^{-1} \Delta x_{t} \nonumber
\end{align}
where $\bar P_t$ is defined as in \eqref{Ppred}. Finally, using $P_t = (H_t^T R_t^{-1} H_t + \bar{P}_t^{-1})^{-1}  = (I - K_t H_t) \bar{P}_t$, we can show that:
\begin{align}
 \max_{x_t} \ \ &  \mu \frac{1}{2} (x_{t} - \bar x_{t})^T V^{xx}_{t} (x_{t} - \bar x_{t}) + \mu  (x_{t} - \bar x_{t})^T v_t^x   \nonumber \\
    &- \frac{1}{2} (x_t -\hat x_t - \hat \mu_t)^T P_t ^{-1} (x_t - \hat x_t - \hat \mu_t)   
\end{align}
where $P_t,  \hat \mu_t$ are defined as in \eqref{Pt1}, \eqref{mut1}.



\bibliographystyle{IEEEtran}
\footnotesize

\bibliography{references}

\end{document}